\documentclass[10pt,twocolumn,letterpaper]{article}

\usepackage[pagenumbers]{cvpr} 

\usepackage{graphicx}
\usepackage{amsmath}
\usepackage{amssymb}
\usepackage{booktabs}
\usepackage{array}
\usepackage[T1]{fontenc}

\newcolumntype{P}[1]{>{\centering\arraybackslash}p{#1}}

\usepackage[pagebackref,breaklinks,colorlinks]{hyperref}

\usepackage[capitalize]{cleveref}
\crefname{section}{Sec.}{Secs.}
\Crefname{section}{Section}{Sections}
\Crefname{table}{Table}{Tables}
\crefname{table}{Tab.}{Tabs.}

\def\cvprPaperID{9309} 
\def\confName{CVPR}
\def\confYear{2022}

\begin{document}

\title{The Introspective Agent: \\ Interdependence of Strategy, Physiology, and Sensing for Embodied Agents}

\author{Sarah Pratt\thanks{Correspondence to \texttt{spratt3@uw.edu}.}\\
University of Washington\\
\and
Luca Weihs\\
Allen Institute for AI\\
\and
Ali Farhadi\\
University of Washington\\}


\maketitle

\begin{abstract}


The last few years have witnessed substantial progress in the field of embodied AI where artificial agents, mirroring biological counterparts, are now able to learn from interaction to accomplish complex tasks. Despite this success, biological organisms still hold one large advantage over these simulated agents: adaptation. While both living and simulated agents make decisions to achieve goals (strategy), biological organisms have evolved to understand their environment (sensing) and respond to it (physiology). The net gain of these factors depends on the environment, and organisms have adapted accordingly. For example, in a low vision aquatic environment \cite{nilsson2014computational} some fish have evolved specific neurons which offer a predictable, but incredibly rapid, strategy to escape from predators \cite{Catania11183}. Mammals have lost these reactive systems, but they have a much larger fields of view and brain circuitry capable of understanding many future possibilities \cite{maciver2017massive}. While traditional embodied agents manipulate an environment to best achieve a goal, we argue for an introspective agent, which considers its own abilities in the context of its environment. We show that different environments yield vastly different optimal designs, and increasing long-term planning is often far less beneficial than other improvements, such as increased physical ability. We present these findings to broaden the definition of improvement in embodied AI passed increasingly complex models. Just as in nature, we hope to reframe strategy as one tool, among many, to succeed in an environment. Code is available at: \url{https://github.com/sarahpratt/introspective}.

\end{abstract}

\section{Introduction}

\begin{figure}[t!]
  \centering
  \includegraphics[scale=0.37]{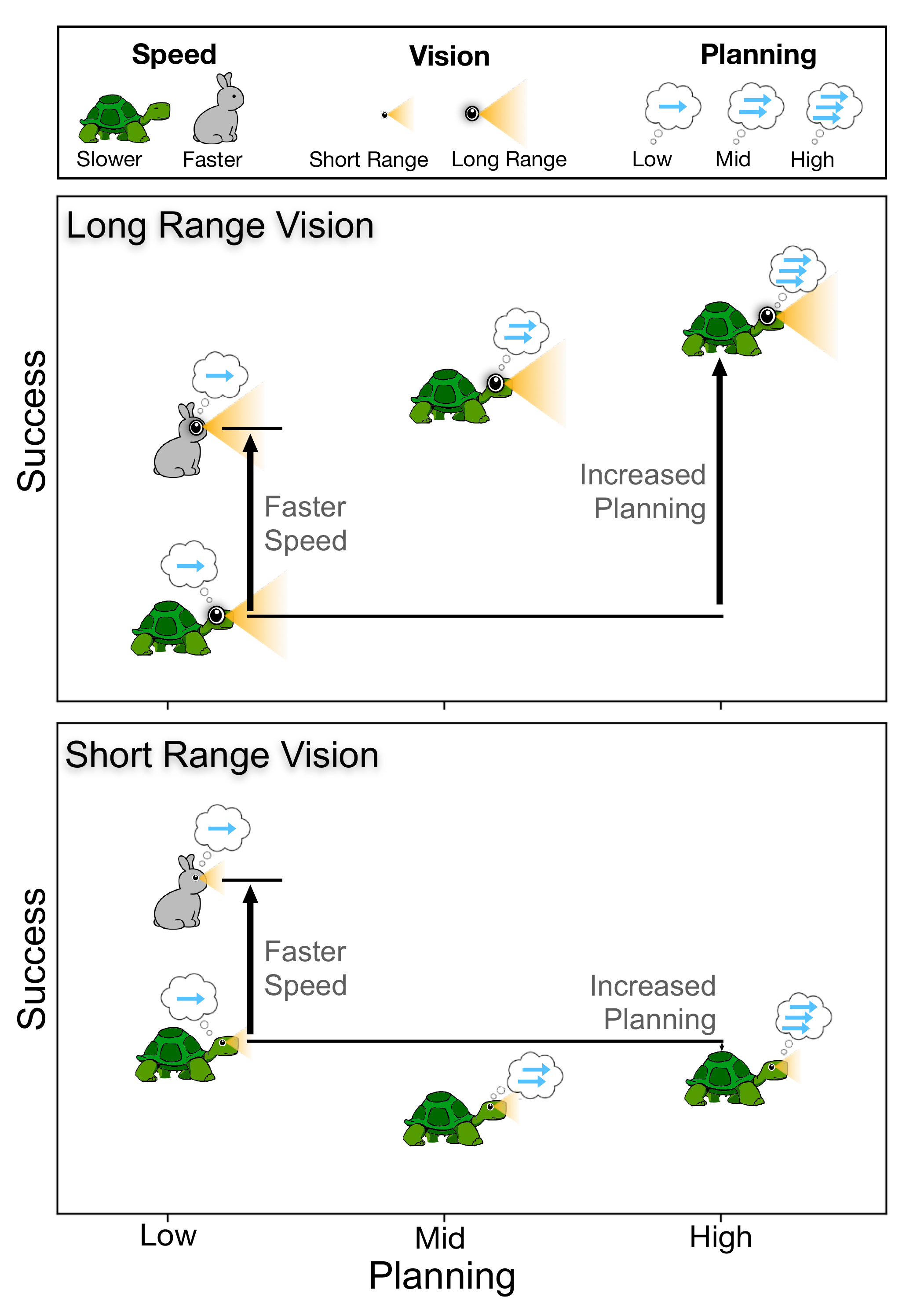}
   \caption{\textbf{The performance of an agent under different design settings.} This demonstrates the interdependence of strategic ability (planning), physiological ability (speed), and sensory ability (visual range), as the improvement from increasing the planning ability versus increasing the speed depends on the value of the visual range.}
   \label{fig:teaser}
\end{figure}

The paradigm of embodied artificial intelligence has expanded the realm of training possibilities for machine learning systems. Agents are able to take an active role in learning, deciding both what do next and what to see next. It has created a relationship between a machine learning model and its environment where, just as in biological environments, agents navigate and manipulate their environment to complete their objective \cite{pfeifer2006body}. 

However, in biological systems the relationship between an organism and its habitat is a dialogue. Just as an agent changes its environment to suit its needs, it too must adapt based on the particulars of its surrounding. There are many examples of this adaptation, from finches evolving their beaks to more efficiently obtain food \cite{darwin2004origin}, to animals changing their coloration based on the season to better avoid predation \cite{jones2018adaptive}. In fact, recent work has hypothesized that the very ability to create complex future plans is a reaction to the changing visibility of the environment as animals moved from the sea to land \cite{maciver2017massive}. Thus, intelligence, the singular goal of much of machine learning research, was not a sweeping solution to all of nature's challenges, but an adaptation that became useful after a change in circumstance. 

In contrast to this evolutionary processes, the relationship between agents and their environments in a conventional simulated settings is entirely one-sided. Agents are hand-designed and then placed into an environment, where they are evaluated based on their ability to overcome challenges in the environment. All improvements over previous methods must be in the agents ability to manipulate the environment more effectively than its successors. This work examines a more introspective agent, which not only considers how to improve its interaction with the environment, but how to improve itself. We study the effect on performance as agents vary across three major factors: strategy, physiology, and sensing. 

Preserving our initial biological inspiration, we examine the interplay of these factors in a simulated predator-prey task. We instantiate a predator and prey in a simulated environment with random obstacles. At the beginning of training, both agents are assigned a specified visual range (sensing), speed (physiology), and planning ability (strategy). We then jointly train both agents using proximal policy optimization (PPO), a reward based reinforcement learning algorithm. The predator receives a reward for capturing the prey, while the prey receives a reward for evading the predator. Thus they are able to learn a policy which is conditioned on their design as well as on the design and policy of their adversary. \\

\noindent Our primary contributions are summarized as follows:

\begin{enumerate}
    \item \textbf{We demonstrate the interdependence of strategic, physiological, and sensory ability in a simulated predator-prey environment.} We examine the relationship between each pair of attributes. Additionally we demonstrate that it is crucial to examine all three when attempting to improve the performance of an agent, as a change in one will effect the optimal value of the others, as visualized in Figure \ref{fig:teaser}.

    \item \textbf{We empirically validate the theory that complex planning emerged as a reaction to the increased view afforded by life on land versus sea, as proposed by MacIver \etal \cite{maciver2017massive}.} We show this by examining these factors through the unified currency of energy. We let capturing prey correspond to a gain in energy and increased planning, speed, or vision correspond to a loss in energy. We find that the cost of these factors has a large effect on the optimal solution, corresponding to the differing values of these factors in aquatic and terrestrial life. We reproduce aquatic organisms with optimal performance at short range vision and low planning ability, as well as mammal and birds which have optimal performance at long range vision and high planning ability.  Additionally, we explore an experimental scheme corresponding to the highly intelligent life that has returned to sea after evolving to live on land, such as whales or dolphins. We find that high planning is also useful in this scenario, where agents are trained with long range vision and later evaluated in a short range vision setting.  

    \item \textbf{We review how to incorporate these findings into  the development of embodied AI.} We discuss when conventional methods of improvement may benefit from an examination of factors other than strategic ability. We argue this is especially crucial when there is a constraint on the resources of a system.

\end{enumerate}

\section{Related Work}

\noindent \textbf{Sensing, Physiology, and Strategy in Biology.} There are a number of works which examine the relationship between these systems in biology. Observations of both fish \cite{maciver2010energy} and crabs \cite{weissburg2003fluid} show that these animals will adjust their body position to advantage either efficient movement or sensing. Niven \etal \cite{niven2008energy} explore the trade-off between energy usage and sensory ability. MacIver \etal \cite{maciver2017massive} use fossil records to create a timeline of increased visual abilities, increased planning abilities, and the progression of sea life to land.

Additionally, there have been efforts to computationally model these observations to better understand their dynamics. Recently, Mugan \etal \cite{mugan2020spatial} formulated predator-prey agents as a Monte-Carlo tree search to discover environments under which planning is most useful. They find tree search outperforms a sampled path most significantly when the agents have high visual ability and the environment is complex. Gurkan \etal \cite{gurkan2019effects} use a similar Monte-Carlo based approach and find that a larger sensory range is much more beneficial in simulated environments which more closely resemble land. These works demonstrate the importance of examining strategy, sensing, and physiology together jointly in biology and provide a strong motivation for incorporating these factors into embodied agents.

\noindent \textbf{Reinforcement Learning for Competitive Gameplay.} Reinforcement learning has been used extensively to study behavior in competitive and multi-agent environments \cite{busoniu2008comprehensive}. Markov games are an especially popular framework for this area of study \cite{littman1994markov}. There has been great success is a number of single and multi-player games such as Go \cite{Silver2016MasteringTG}, VizDoom \cite{lample2017playing}, Atari Breakout \cite{mnih2015human}, Dota \cite{berner2019dota}, and Starcraft \cite{vinyals2019grandmaster}.

Additionally, a number of works have specifically addressed the problem of navigating to a specific target in an embodied setting. These works address both hidden targets \cite{weihs2019learning} as well as mobile targets \cite{wang2020cooperative, baker2019emergent}. While these works provide a variety of successful examples of competitive reinforcement learning for both single and multi-agent scenarios, for all of these works, the agent's design is always fixed. In contrast, this work examines a number of potential designs to understand the optimal adaptation across multiple dimensions.

\noindent \textbf{Learning Physiology.} A number of fields have examined methods to adapt aspects of an agent's physiology. One such field is robotics \cite{auerbach2014robogen, mintchev2016adaptive}. For many of the works which adapt morphology in robotics, designs are first established in simulation and then later fabricated and tested in a real-world environment \cite{whitman2020modular, lipson2000automatic, cheney2014unshackling, sims1994}. Learning physiology in conjunction with policy has also been used in the fields of evolutionary algorithms \cite{bongard2014morphology}, deep learning  \cite{ha2020fit2form}, and deep reinforcement learning \cite{ luck2020data, yuan2021transform2act, schaff2019jointly} which adapt simulated agents to succeed in entirely simulated tasks.

Ha \cite{ha2019reinforcement} uses deep reinforcement learning to adapt a 2D agent. In addition to learning a policy, the algorithm learns the design of the agent's legs. The agent is able to learn an effective design and strategy which outperforms the hand-designed agent. Gupta \etal \cite{gupta2021embodied} extends this to a 3D environment with many more design possibilities. They find that different designs and strategies emerge for different environments and the design of the agent determines the time it takes to learn effective strategies. These works are a valuable step towards pushing embodied AI passed policy learning and demonstrate the importance of agent design. However, these works only allow agents to learn the action space, and do not allow agents to adapt how they perceive their environment.

\begin{figure*}[t!]
  \centering
  \includegraphics[scale=.28]{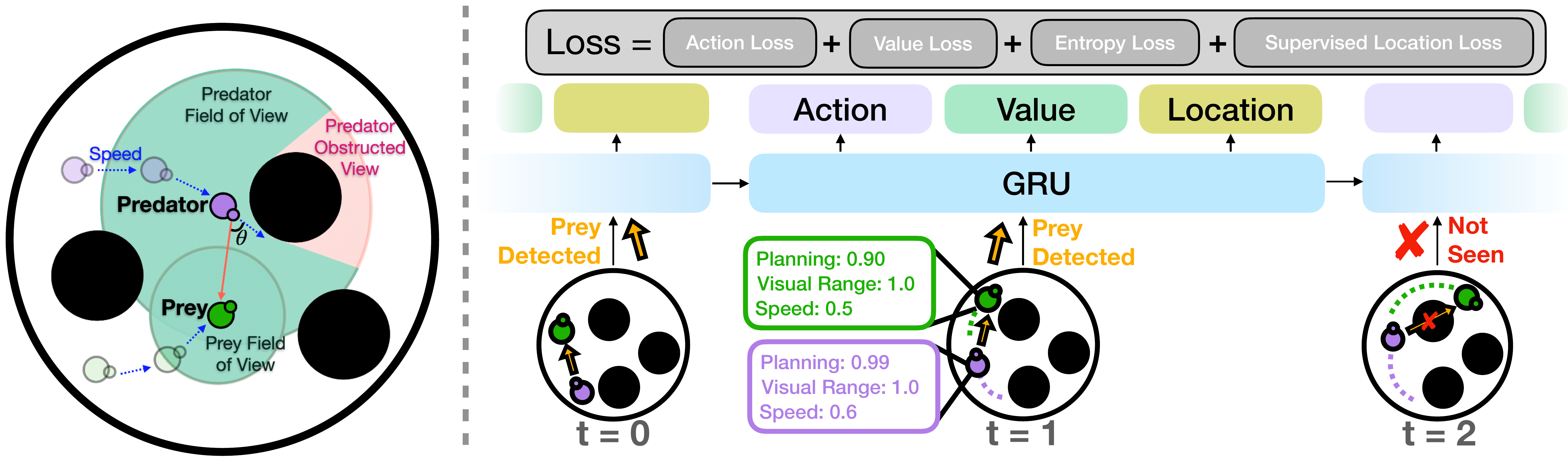}
   \caption{\textbf{Schematic of environment and model.} Each agent is modelled with a GRU, which outputs an action distribution, value prediction, and a prediction of its adversary's location. Here we visualize just the model for the predator as the prey's is identical.}
   \vspace*{-0.3cm}
   \label{fig:model}
\end{figure*}

\noindent \textbf{Learning To Sense.}
Adapting an agent to sense the environment is a less investigated topic, but there have been a few promising efforts in this direction. Early work by Sims \cite{sims1994, sims1994evolvingdim} does examine the combination of sensing and physiology, but does not include any method of learning or optimization other than random permutation. This is an important contribution as it allows agents to develop both of these factors, but agents are not able to adjust their policy to best match their physical and sensory abilities. 

Work in robotics has examined sensory ability and strategy with adaptable camera positions to allow quadrotors to navigate environments of variable complexity \cite{sanket2021morpheyes}. Additionally, Tseng \etal \cite{tseng2021differentiable} jointly optimize the design of a camera and the image post-processing based on the downstream application. Other work allows an agent to maneuver both its gripper and its camera to allow control over the information gathered by the manipulator \cite{cheng2018reinforcement}. Finally there is work inspired by selective attention which allows networks to select a small portion of an image to view at each time step \cite{tang2020neuroevolution}. Though this is not directly analogous to evolving sensors, it demonstrates the value and effectiveness of adaptable visual input.

\section{Methods}
The overall methodology of this work is to model our agents each as a separate deep neural network. Additionally, each agent has parameters which define their strategic, physiological, and sensory abilities. We then train these agents jointly against each other so that they may learn a policy conditioned on their design and the design of their adversary.

\subsection{Task}
To study the interplay between sensing, strategy, and physiology, we select a task where an agent's performance is deeply impacted by these factors.
As we have evidence that these components impact animals' performance in hunting and predator evasion \cite{maciver2017massive}, we select a simulated predator-prey scenario as our task. We model our environment after that in Lowe \etal~\cite{lowe2017multi}, with some modifications detailed below.\\

\noindent \textbf{Environment:} A visualization of our environment can be found in Figure \ref{fig:model}. The environment itself is a circular arena. At the perimeter of the circle is a wall through which agents cannot pass. Additionally, the environment contains obstacles. In all of the experiments in this work, there are three obstacles which are instantiated in random locations within the circular area. Obstacles act as barriers which halt agent movement and occlude other agents from view. \\ 

\noindent \textbf{Agents:} In our experiments there is one predator, whose goal is to capture the prey, and one prey, whose goal is to evade the predator. In each instance of the environment, the predator and prey are instantiated at random locations.

Beyond their 2D location within the environment, agents also have a heading direction. Thus, like many animals, they cannot move easily in any direction, but must rotate to face the direction they plan to move. Each agent is able to take one of four actions at any time step: (1) move forward by $n$, (2) rotate $x$ degrees right and move forward by $n$, (3) rotate $x$ degrees left and move forward  by $n$, or (4) do not move. The degrees rotated by the agents scale linearly with the speed of the agent, so that a faster agent is also able to rotate more rapidly.

\subsection{Model}
\noindent \textbf{Architecture} The primary element of the network architecture for each agent is a Gated Recurrent Unit (GRU) \cite{chung2014empirical}, which enables the agents to implement planning as information from the previous time step can be passed to the next one. 
As input the GRU receives the current observation of the agent, described in Section \ref{sec:training}, as well as an embedding of the previous action taken by the agent. 

We train this architecture with PPO, \cite{schulman2017proximal}, an Actor-Critic methodology. As this name implies, the model contains both an actor, which outputs the distribution of actions that the agent will take, and a critic, which outputs the predicted value of that action distribution. In our implementation, both the actor and the critic consist of three-layer MLP, with a $tanh$ non-linearity. Each of these heads receives the hidden state of the GRU as input. Additionally we found that, for training stability, it is helpful to include an auxiliary loss that requires an agent to predict the location of its adversary. For this reason, our model also contains a location prediction head. This head outputs a prediction of the relative position of the agent's adversary. The GRU and linear layers all have a width of 512.

\subsection{Training Details}
\label{sec:training}

\noindent \textbf{Algorithm:} As previously mentioned, we use proximal policy optimization (PPO) \cite{schulman2017proximal} in order to optimize our model. We train our model using roll-outs of 50 consecutive steps and a batch size of 64. While it is common to perform more than one gradient update at each gradient step, we perform only one gradient update for each of these roll-outs which we found to improve the stability of training. We train our model for $10{,}000$ gradient updates, corresponding to $500{,}000$ total steps in the environment. We use an Adam optimizer \cite{kingma2014adam} with an initial learning rate of of $7\cdot 10^{-4}$, $\beta=(0.9, 0.999)$, and $\epsilon=1\cdot 10^{-5}$. We linearly decay the learning rate to zero throughout training.\\

\noindent \textbf{Loss:} To train our model, we use a weighted sum of four losses. Three of these are as described in \cite{schulman2017proximal}. We use an action loss which optimizes the reward of the actions produced by the actor, a value loss which optimizes the value predictions made by the critic, and an entropy loss which helps prevent early policy collapse. 

Additionally, we use a supervised loss on the location prediction of the model. This is a cross entropy loss on the location-prediction head of the model, which outputs a singular location corresponding to 144 possible different relative locations (8 possible distances $*$ 18 possible angles). We find this supervised loss is crucial for training. \\

\noindent \textbf{Rewards:} The prey receives a reward of $-5$ if it is captured by the predator and otherwise receives a reward of 0. The predator receives a reward of $+5$ for capturing the prey. If it has not captured the prey by time step 400, it received a reward of $-1$. In either case, the environment is reset with random agent and obstacle locations. \\

\noindent \textbf{Observations:} At each time step, the agent (predator or prey) receives an observation containing the following information:
\begin{itemize}
  \item \textbf{Adversary location}: If the adversary is visible to the agent (within the field of view and not blocked by an obstacle), the agent receives (1) the relative angle between its current heading direction and the adversary's location ($\theta$ in Figure \ref{fig:model}), (2) the agent's distance from the adversary, (3) the current heading direction of the adversary. If the adversary is not visible, then the agent receives $[-1, -1, -1]$ for these three values.
  \item \textbf{Obstacle locations:} At each time step, the agent receives (1) the relative angle between each obstacle and the agent's current direction (2) the distance from the center of each obstacle to the agent. This information is always given to the agent regardless of its current visual range. 
  \item \textbf{Self location:} The agent is also told its own current location in the environment given by (1) the global angle of the agent, i.e. the angle of the line which passes through the center of the environment and the center of the agent, (2) the global distance of the agent, i.e. the distance from the agent to the center of the environment and (3) the current heading direction of the agent. Again, this information is always given to the agent regardless of its current visual range.
\end{itemize}

\subsection{Strategy, Physiology, and Sensing Parameters}
We now discuss how we parameterize the three factors we wish to examine: physiological, sensory, and strategic ability. \\

\noindent \textbf{Physiological Ability.} In nature, there are countless metrics of an organisms physiological ability from strength, to dexterity, to stealth. For simplicity, we use a single parameter which linearly scales the speed and the turning agility at each time step. As our environment is simulated using discrete time steps, this corresponds to the size of the step which an agent can take at each discrete time as well as the number of degrees it can turn. The prey speed is set to 0.50. We select the average predator speed based on the lowest estimate of a typical speed ratio found in terrestrial life \cite{mugan2020spatial}. In our work we consider five different possible speeds for the predator: 0.50 (very slow), 0.55 (slow), 0.6 (average speed), 0.65 (fast), 0.70 (very fast).  \\

\noindent \textbf{Sensory Ability.} Like physiology, there are many ways that organisms have adapted to sense their surroundings. In this work, we define the sensory ability of the agent as the area of the field of view. For simplicity, we define the field of view as a circular region around the agent, as visualized in Figure \ref{fig:model}. We consider three areas of view: 0.3 (short range vision), 1.0 (medium range vision), and the entire region (long range vision). In all these cases, an agent can see its adversary if is within the field of view and not occluded by an obstacle. The relative size of these fields of view, the environment, and the obstacles are visualized in supplementary material.\\

\noindent \textbf{Strategic.} To adjust the strategic abilities of the agents, we take advantage of the way the action loss is computed. In order to encourage positive actions which do not immediately yield a reward, the action loss uses the $\gamma$ discounted cumulative reward which takes the following form:
$$A_t = R_t + \gamma  R_{t+1} + \gamma^2 R_{t+2} + \gamma^3  R_{t+3} + ...$$
where the $R_t$ is the reward obtained at time $t$ and $\gamma$ is a value from 0 to 1. Thus, as $\gamma$ decreases, so does the contribution of future rewards. As an illustration of this effect, a reward which is obtained 50 steps in the future will still retain ~60$\%$ of its initial value if $\gamma = 0.99$, but will retain less than 3$\%$ of its initial value if $\gamma = 0.93$. A model with a smaller $\gamma$ value will thus be incentivized to take actions that maximize short-term gains and vice versa. 
We therefore utilize this parameter as a measure of long-term planning ability. We consider three values of $\gamma$: 0.90 (low planning), 0.93 (mid planning), and 0.99 (high planning). \\

In our experiments we hold the above values constant for the prey and vary them for the predator. We do this to better examine the interaction between these factors, without the added dimension of variable prey. Note that the prey still learns to adapt its policy to optimize its performance against the predator, we simply do not also consider, \eg, varying the prey's speed.

\section{Experiments}

\noindent \textbf{Evaluation Metric.} For many of the following experiments, we are concerned, not just with the catch success of the predator, but also with the energy cost of the design of the predator. We use an evaluation metric to reflect both these concerns. Therefore rather than using the typical metric of total successful runs, we hold the number of steps constant in an effort to hold the energy spent constant, and instead examine the total number of prey captured in $10^4$ steps. Additionally, to avoid outlier episodes where the predator is unable to capture the prey at all, we reset the locations of the agents and the obstacles if the prey has not captured the predator in 400 steps, just as we do during training. This allows us to understand the true energy gain of the predator while keeping the energetic cost associated with the number of steps constant. Thus when we examine the energy costs, we can examine only the costs associated with different model designs, as the cost associated with the number of step is equal for all experiments. \\

\begin{figure}[t!]
  \centering
  \includegraphics[scale=.35]{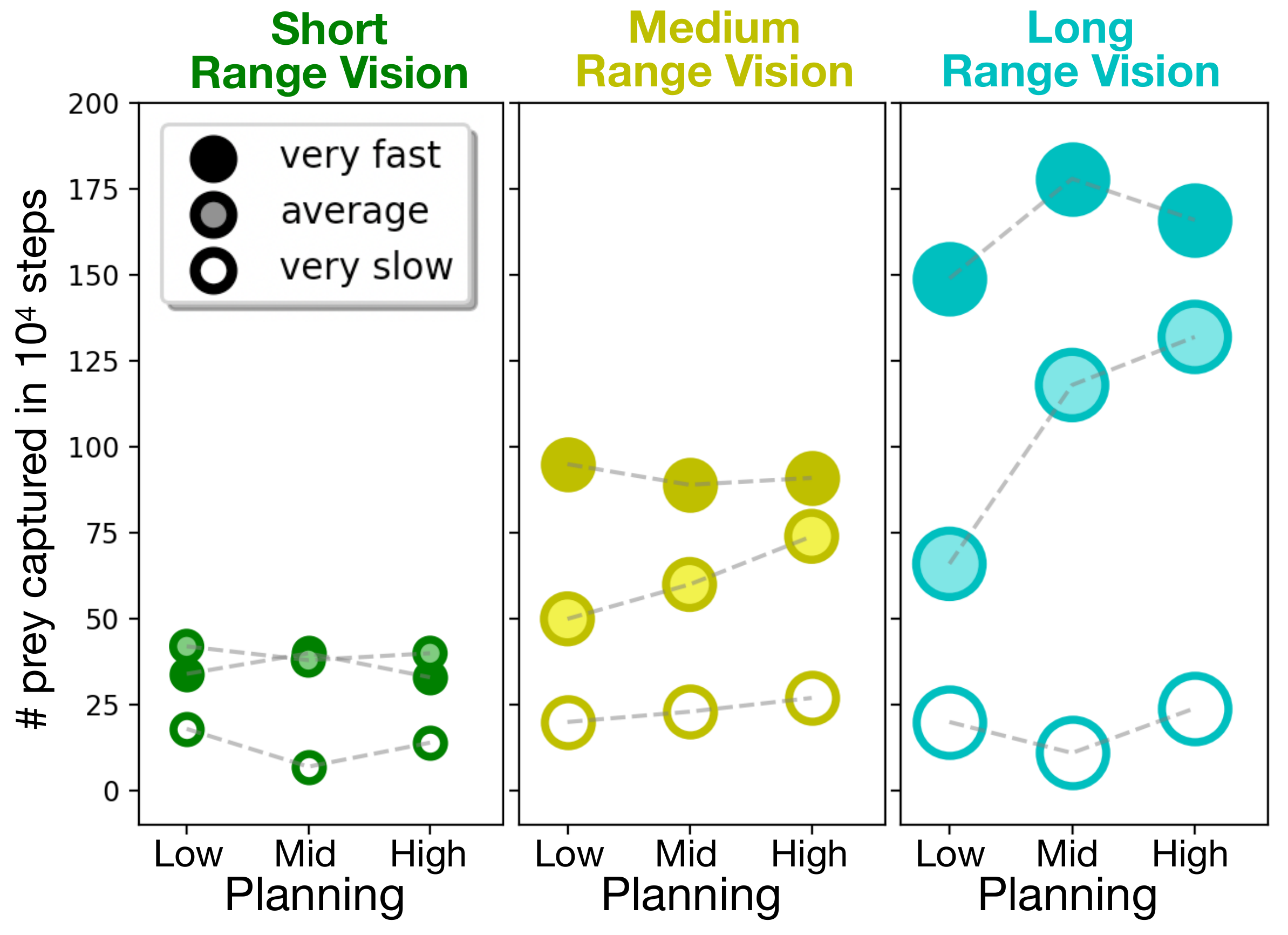}
   \caption{\textbf{Performance of predator agent across different sensory (vision), physiological (speed), and strategic (planning) abilities.} Each point represents a predator trained with a different combination of these values. Each predator is evaluated against a prey which has been trained jointly so it is best able to evade a predator with the specified design.}
   \label{fig:three}
\end{figure}

\subsection{Results}

\subsubsection{Effects of Strategy, Physiology, and Sensing}
\label{sec:effects}
We first examine the relationships that appear between strategy, physiology, and sensing, and how the impact of adjusting one can drastically change based on the values of the others.\\

\noindent \textbf{Strategy and Sensing.}
In this setting, we examine the effect of different planning abilities across different sensory abilities for a fixed speed value. As demonstrated in Table \ref{tab:example}, the value of higher planning varies drastically depending on the visual range. With short range vision, planning ability makes no difference, the high planning predator catches the prey 0.952 times as much as the low planning predator. With medium range vision, planning begins to make a difference, with the high planning predator catching the prey 1.48 times as much as the low planning predator. With long range vision, planning makes a large difference. The high planning predator catches the prey 2 times as much as the low planning predator. 

We can see this trend visually by examining the slope of the lines in Figure \ref{fig:three}. When considering an average speed (represented by the semi-transparent points), we see that as the visual area increases, so does the slope associated with an increase in planning ability. \\

\noindent \textbf{Strategy and Physiology.} In this setting, we examine the relationship between the planning ability and the speed of the predator for a fixed visual ability. Again, we see that it is impossible to understand the effect of one of these values without considering the value of the other. As shown by Figure \ref{fig:three}, the importance of planning is the most pronounced at a moderate speed. If the speed is too high or low, increased planning no longer leads to increased performance. 

This is represented visually by examining the slope of the three lines within a single panel in Figure \ref{fig:three}. While the first panel shows no positive correlation with planning in any setting, the second and third show a pronounced positive correlation only in the average speed setting (represented by the semi-transparent points). A very low or high speed (represented by the fully transparent and opaque points respectively) show little or no increase in performance with an increase in planning ability.

We conjecture that this is because, at very high speed, complex planning is not necessary as the predator is able to perform optimally using physical ability alone. Conversely, at very low speeds, no amount of planning ability can overcome the lack of physical ability.
\\

\begin{figure}
  \centering
  \includegraphics[scale=.35]{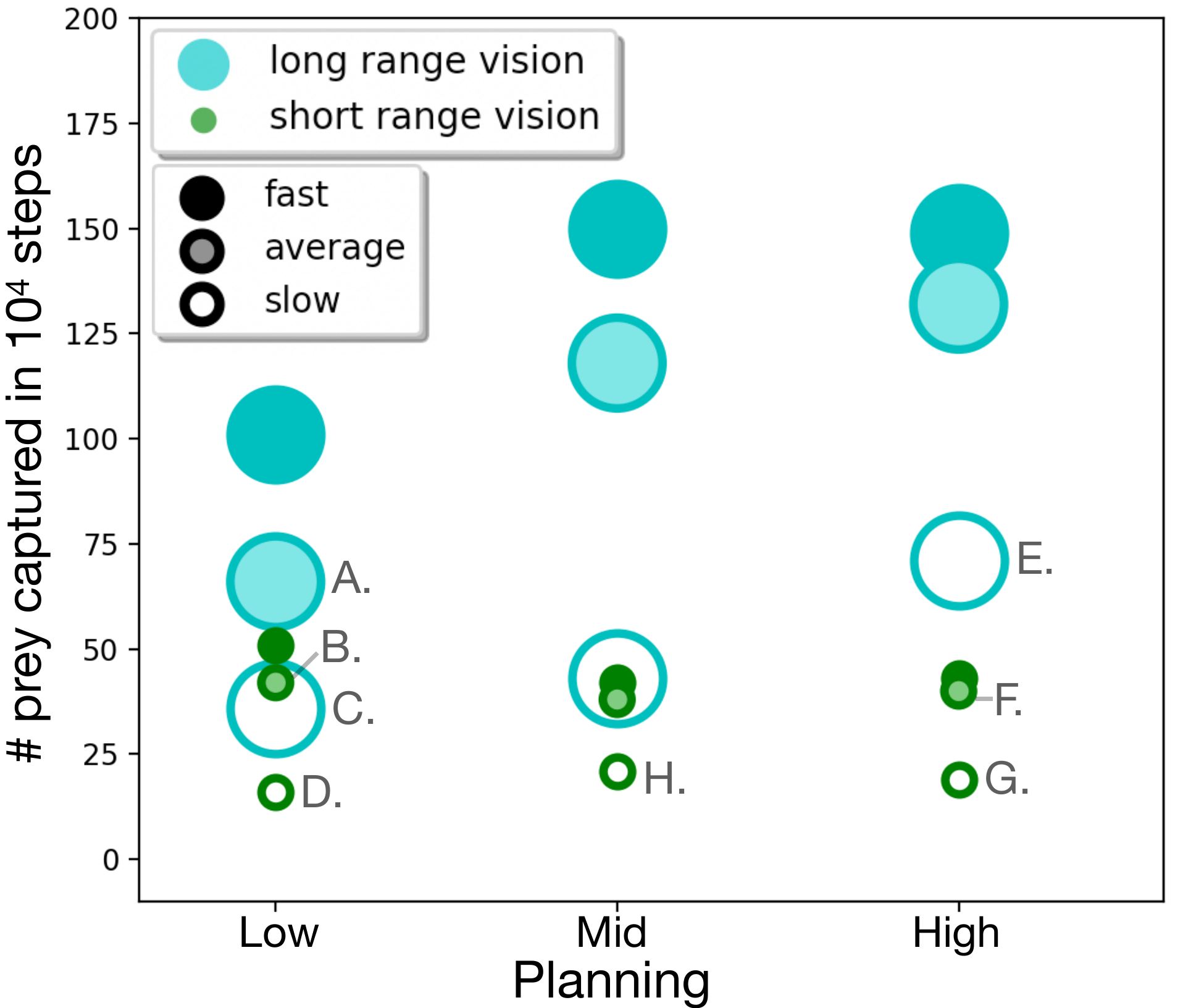}
   \caption{\textbf{Most effective improvement of predator depends on current value of all three factors.} We must examine all factors to know if a predator will see the largest gain from faster speed, longer vision or higher planning. More details under \textbf{Strategy, Physiology, and Sensing} in section \ref{sec:effects}.}
   \label{fig:all}
\end{figure}

\begin{table}
\begin{tabular}{ |p{1cm}||P{0.5cm}|P{0.5cm}|P{0.5cm}|P{1.3cm}|P{1.2cm}|  }
 \hline
  & \multicolumn{3}{|c|}{Planning}&\multicolumn{2}{|c|}{Diff(high, low)} \\
 \hline
  Vision & low & mid & high & Abs Diff & Rel Diff\\
 \hline
 short & 42 & 38 & 40 & 2 & 0.952 \\
 medium & 50 & 60 & 74 & 24 & 1.480 \\
 long & 66 & 118 & 132 & 66 & 2.000 \\
 \hline
 
\end{tabular}
\caption{\textbf{Values for visual range vs. planning at average speed}}
\vspace*{-0.3cm}
\label{tab:example}
\end{table}

\begin{figure*}[t!]
  \centering
  \includegraphics[scale=.32]{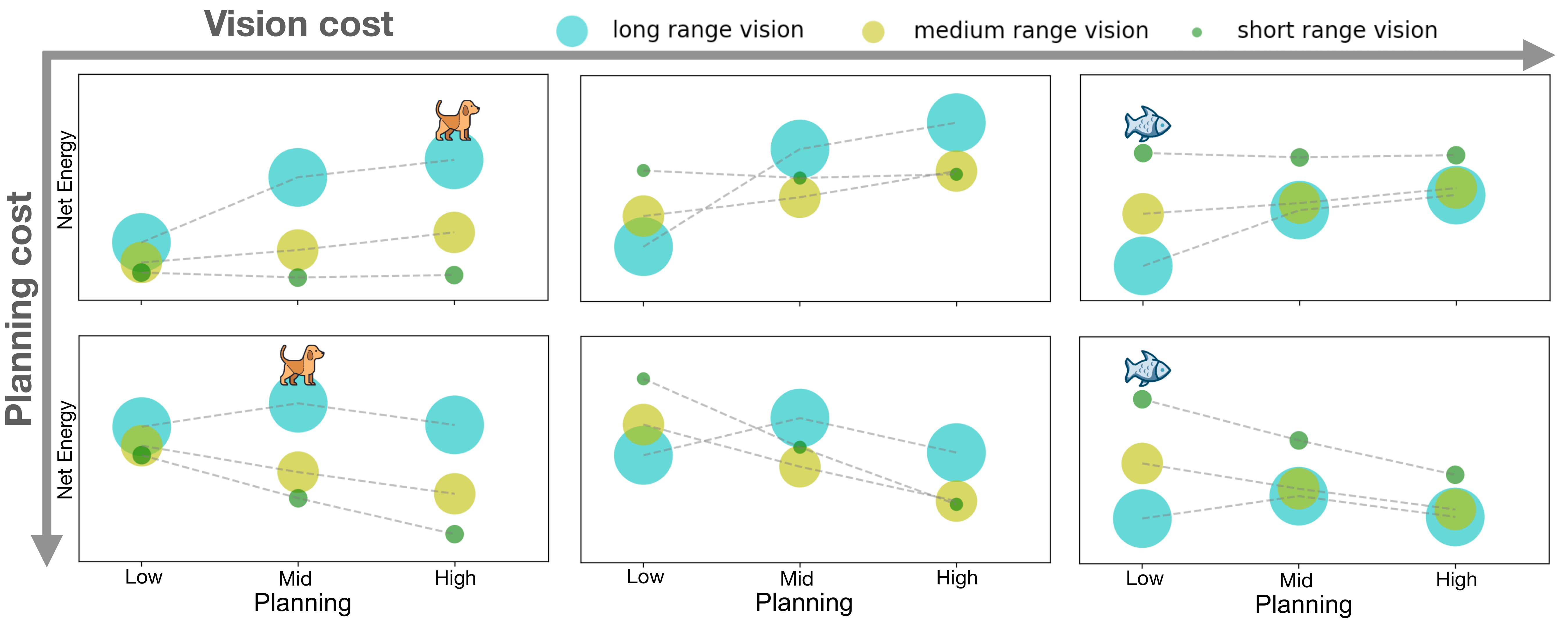}
   \caption{\textbf{Net energy gain at different visual and planning abilities considering different energy costs of these abilities.} We base the cost of increased ability on $\sigma$, or the standard deviation of the gross energy gain of all predator configurations. For visual range we consider costs of 0, $\sigma$, and $2 \cdot \sigma$ for each increase in ability. For planning ability, we consider costs of 0 and $\sigma$ for each increase in ability. The dog and fish icons indicate the optimal setting under those given costs, which we find reproduce the abilities of terrestrial and aquatic life, respectively \cite{maciver2017massive}}
   \vspace*{-0.3cm}
   \label{fig:cost}
\end{figure*}

\noindent \textbf{Sensing and Physiology.} The relationship between sensing an physiology is also apparent in Figure \ref{fig:three}. As we increase the sensory ability of the predator, the difference between high speed and low speeds becomes more exaggerated. Fixing the planning ability at high, the fastest speed performs 2.36 times better than the slowest speed at the short visual range, 3.37 times at the medium visual range, and 6.91 times better at the long visual range. \\

\noindent \textbf{Strategy, Physiology, and Sensing. } Finally, we are able to make observations about how all three of these factors interact.  We are able to observe cases where we must consider the entire system in order to know what changes will have the biggest effect on performance.  

The first case we consider is the setting where our system has poor physical and visual ability (points D, H, and G in Figure \ref{fig:all}). In order to know what change to this system will lead to the biggest improvement, one must consider the third factor -- planning ability. At a high planning ability (G), an increase in visual ability (E) leads to a greater improvement than an increase in speed (F). At a low planning ability (D), an increase in visual ability (C) leads to a worse improvement than an increase in speed (B).

The second setting where we must consider all factors is when our system has poor physical and planning ability (points C and D). In this case we must consider the value of the sensory ability in order to understand how to adjust the other two. At a high sensory ability (C), an increase in planning ability (E) leads to a slightly greater improvement than an increase in speed (A). At a low sensory ability (D), an increase in planning ability (G) leads to a worse improvement than an increase in speed (B). Note that these trends hold when examining a speed increase of 0.55 to 0.60 (transparent to semi-transparent) or 0.60 to 0.65 (semi-transparent to opaque).

These results demonstrates that in some cases it is crucial to examine all factors of a system in order to understand how a single factor may impact performance. While sensory, physical, and strategic ability may seem as though they are independent axes all capable to of leading to uncorrelated levels of improvement, this is not the case. The effect one factor can have on performance is deeply tied to the values of the others.\\ 

\subsubsection{Reproducing Biology}

\noindent \textbf{Aquatic Life versus Terrestrial Life.} In addition to understanding the dynamics which emerge between physiological, strategic, and sensory abilities in our simulated environment, we observe patterns which mirror those we see in biology. As previously discussed, work in evolutionary biology has hypothesized that long term planning potentially emerged as a response to the higher visibility of life on land \cite{maciver2017massive}. High sensory ability made this type of intelligence advantageous for catching prey and avoiding predators. Therefore, high planning ability should not be viewed as a "more optimal" strategy, but a strategy which is optimal for a different set of circumstances.

To examine this concept in our framework, we examine the costs and benefits of sensory and strategic abilities through the lens of energy. If there is no associated energetic cost, then selecting the highest proficiency for each ability is a good policy. Having excellent visual and planning abilities is unlikely to disadvantage an organism. However, both of these abilities have been shown to be limited by metabolic energy cost \cite{niven2008energy, aiello1995expensive}. While a higher planning ability may contribute to a more successful predator and therefore a higher gain in energy, it is very energetically costly to run these systems. Additionally, prior work in robotics has shown that under different scenarios, systems will find different optimal policies to minimize energy consumption \cite{fu2021minimizing}.

To understand the optimal strategies which emerge under various environments, we examine the relationship between vision and planning both in terms of the catch success as well as the environmental cost. Though there are many factors at play when considering organisms in different ecosystems, we simplify our examination of aquatic and terrestrial habitats into two assumptions: (1) the same visual range is more costly in water than on land \cite{maciver2017massive} and (2) the same level of planning is equally costly in water and on land.

As we see in Figure \ref{fig:cost}, varying the costs associated with vision creates different optimal strategies. Unsurprisingly, as vision becomes more energetically expensive, the optimal design is to have a lower vision. More surprisingly, as the cost of vision increases, it also becomes more beneficial to have a lower ability to plan, as demonstrated by the right column of Figure \ref{fig:cost}. The low visual and planning abilities of the highest performing agent reproduces the visual and planning ability we see in aquatic life \cite{maciver2017massive}, in an environment designed to mirror the difficulty of long range vision in an ocean environment.  
The left column demonstrated the optimal design when long range vision is inexpensive. As vision does not have a high cost, it is beneficial to have a high visual range. Additionally, in this setting, it is also beneficial to have a high ability to plan. This reproduces the high sensory and planning abilities we see in land animals such as mammals and birds, in an environment designed to mirror the long range vision afforded to these animals.

This directly mirrors the hypothesis put forth by MacIver \etal \cite{maciver2017massive} and furthered by Mugan \etal \cite{mugan2020spatial} which claims that long range vision afforded by life on land led to development of long-term planning. When long range vision is inexpensive, we see long-term planning as the optimal design. However, when it is very difficult to have long range vision, it is best to have minimal planning abilities.\\

\noindent \textbf{Formerly Terrestrial, Aquatic Life.} While there are known instances of limited planning of aquatic life \cite{Catania11183}, there has been significant research into the complex minds of whales and dolphins \cite{simmonds2006into}. These animals provide a unique look into the effect that life on land has on planning ability, as they adapted to life on land before evolving to life underwater once again. 

\begin{figure}[t!]
  \centering
  \includegraphics[scale=.18]{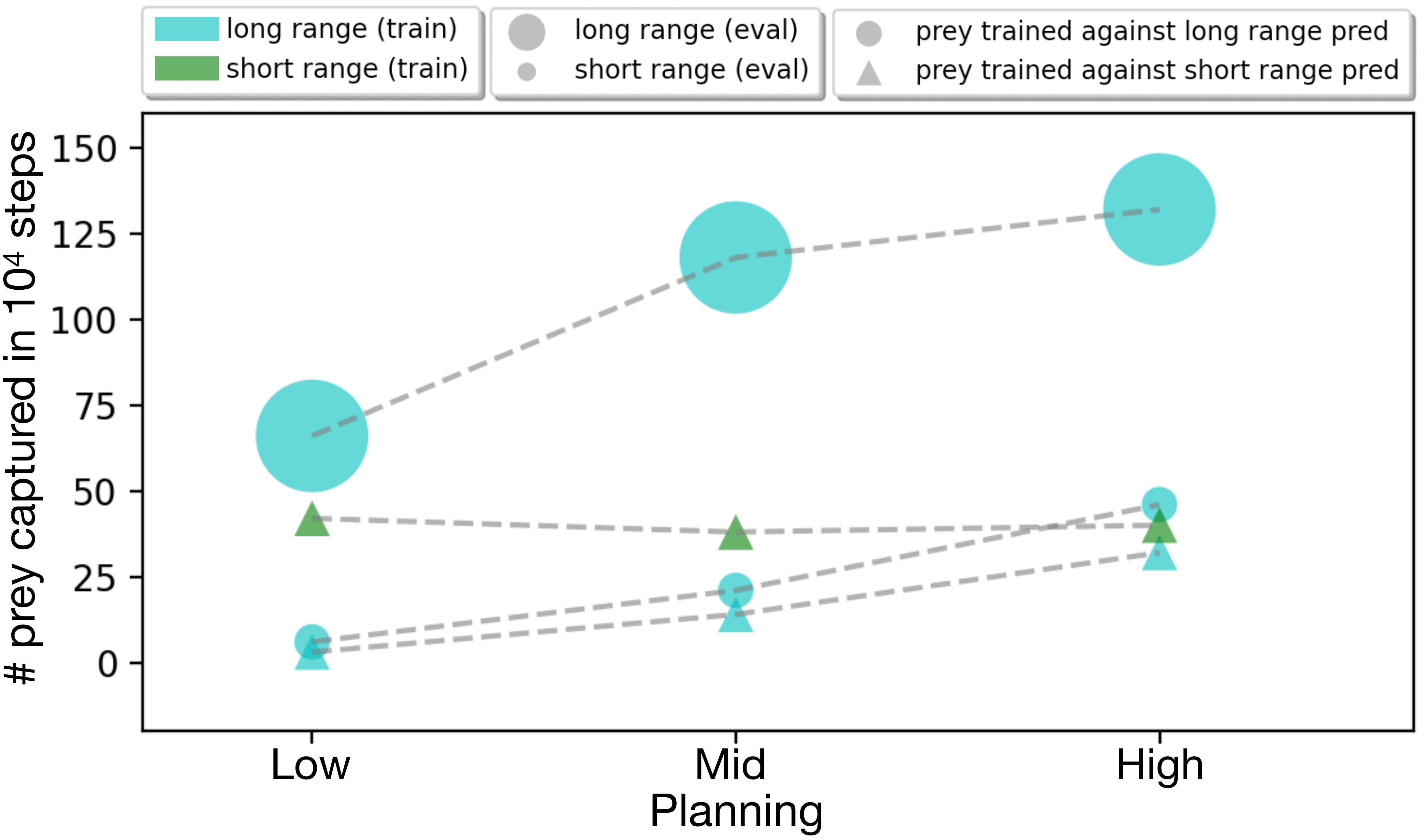}
   \caption{\textbf{Effect of planning when visual range differs between training and evaluation.}}
   \vspace*{-0.5cm}
   \label{fig:whale}
\end{figure}

We model this development, by training a predator in a long range vision setting, and then evaluating it in a short range vision setting. We evaluate against a prey that has been trained to evade a long range vision predator and a prey that has been trained to evade a short range vision predator. In both settings, we find that planning is a useful adaptation as demonstrated in Figure \ref{fig:whale}. This is surprising as planning does not prove to be a useful adaptation if the predator is trained and evaluated in a short range vision environment. This sheds light on instances where complex planning may still be useful, even when high visual abilities are not.

\section{Discussion}
Understanding the dynamics between different attributes is not only important for understanding the natural world, but for making improvements in the world of embodied AI. As we have shown, in the right circumstances improving a models ability to strategize can have a large impact on its success. We have seen many examples of such gains in the realm of embodied AI, with more capable model making huge strides on a variety of difficult tasks.

However, more capable models are only one part of a larger story. Additional attributes of an agent can have a large impact on the benefits of improved models. We have shown how other considerations may impact the effectiveness of models in two ways. 

Firstly, when there exist resource constraints, better models may not be the best way to improve performance. In the context of biology, these constraints are on the metabolic energy of the organisms. In embodied AI or robotics, these constraints may be dictated by the battery of the system or the financial cost of system. As shown by this work, increased computational efforts may not always give the best improvement under these constraints. In some circumstances, larger gains may be obtained by considering the physical or sensory capabilities of the model. 

Secondly, even when no constraints exists, it is important to consider additional factors in order to see the maximum improvement from a better model. As we demonstrated, a model capable of a higher performance did not have a higher accuracy when the speed or sensory abilities of the system were too low. Only when these improved, did the better model lead to an increase in performance.

\section{Limitations and Future Work}
Our primary limitation is our inability to capture the true complexity of biological systems. There are many ways that organisms have adapted to their environment. In order to understand their interplay, we limit our scope to three parameters. There is equal complexity in the environments of these organisms, which, for simplicity, is also limited in this work. We aim to address both of these concerns in future work by examining more complex aspects of strategy, physiology, and sensing as well as more complex environments. However, we acknowledge that no simulated environment will capture the complexity of our world. Additionally, we plan to automate the optimization of these factors to increase agent performance across multiple dimensions in a single training run. We hope this work proves as a useful step to broaden the possibilities of improvement in embodied AI. 

\noindent \textbf{Acknowledgments} We would like to thank Kiana Ehsani, Aaron Walsman, and Mitchell Wortsman for their valuable insight and suggestions over the course of this work.

\typeout{}
{\small
\bibliographystyle{ieee_fullname}
\bibliography{main}
}

\clearpage

\appendix

\begin{figure}
  \includegraphics[scale=0.37]{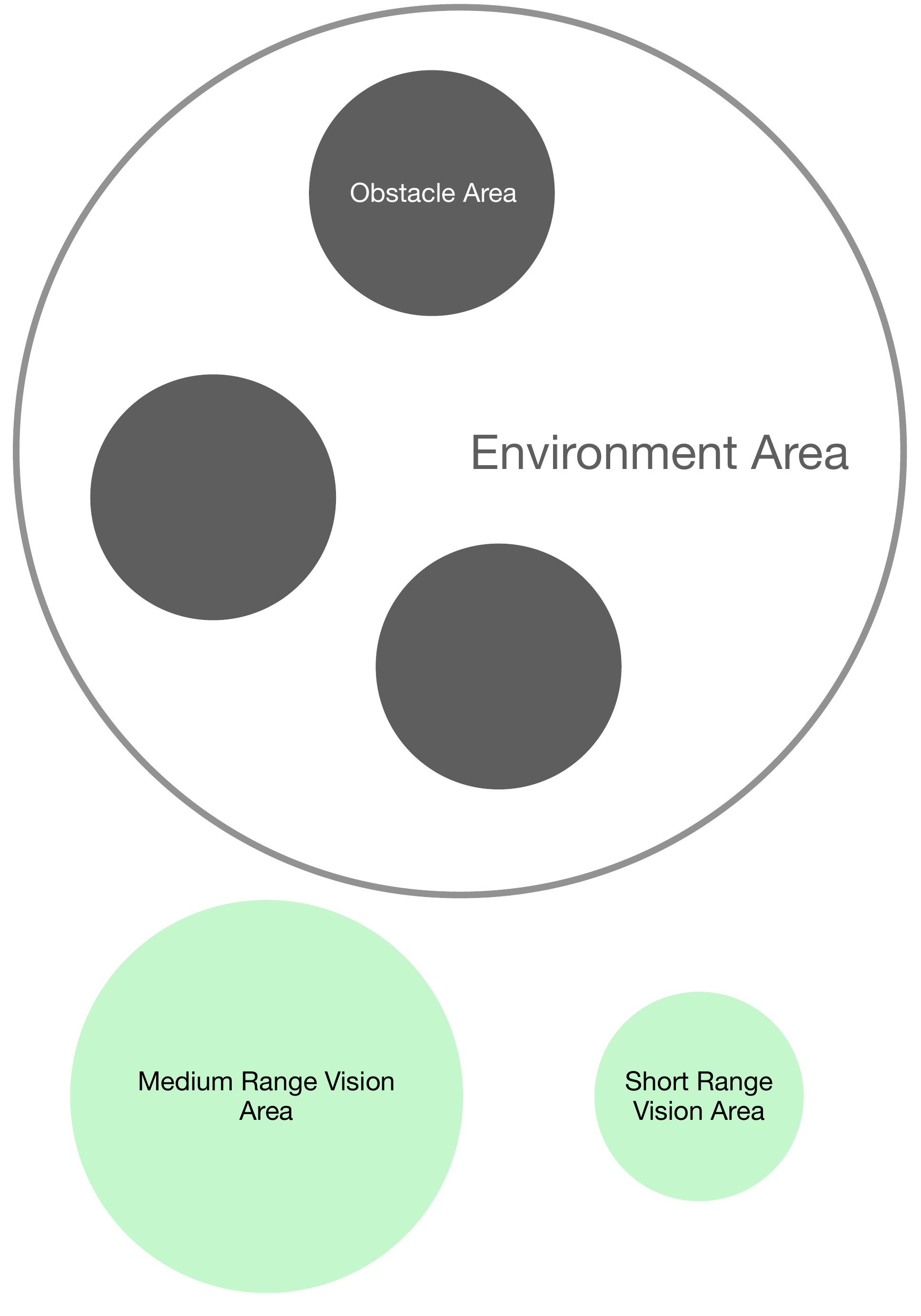}
   \caption{\textbf{Visualization of scale of items in the environment.}. The area of long range vision is not included as it is just the entire environment (excluding occluded areas).}
   \label{fig:scale}
\end{figure}

\section{Additional Environmental Details}
All details of the environment can be found in the code, however we highlight several aspects here.

\subsection{Scale}
As mentioned previously, in this work we investigate three possible sizes of the field of vision: area of 0.3, area of 1.0, and the entire environment. The environment has radius 1.3 (area of 5.3) and the obstacles have radius 0.35 (area of 0.38). The agents themselves have a radius of 0.04 (area of 0.005). There are three obstacles in each environment. The areas of the obstacles, environments and fields of view are visualized in Figure \ref{fig:scale}.

\section{PPO implementation}

We utilize an open source implementation of the PPO \cite{schulman2017proximal} algorithm for our implementation \cite{pytorchrl}, which we modify to include our previously described location prediction head and supervised location loss. Hyperparameters are as listed in Table \ref{tab:hypers}.
Additionally, we include the code used to train, evaluate and visualize the experiments in this work.

\begin{table}
\begin{center}
\begin{tabular}{ |p{3.5cm}||P{2cm}|  }
 \hline
  \multicolumn{2}{|c|}{Hyperparams for PPO} \\
  \hline
  \hline
   Batch Size & 64  \\
  PPO Clipping & 0.2  \\
  Clip value loss & False  \\
  Policy Epochs & 1 \\
  Entropy Coefficient & 0.01  \\
  Value Loss Coefficient & 0.5  \\
  Use GAE & False  \\
  \# Mini Batches & 1 \\
  \# Rollout Steps & 50  \\
  Total Env Steps & 5e4  \\
  Initial LR & 7e-4  \\
  Max grad norm & 0.5  \\
  LR scheduler & Linear decay \\
  Optimizer & Adam \cite{kingma2014adam} \\
 \hline
\end{tabular}
\caption{\textbf{Hyperparameters for PPO}}
\vspace*{-0.3cm}
\label{tab:hypers}
\end{center}
\end{table}

\section{Assets}

We use two pre-existing assets in this work. 

\begin{enumerate}
\item Our environment, a which we modify from Lowe \etal \cite{lowe2017multi} (which itself uses an environment from Mordatch \etal \cite{mordatch2017emergence}).
\item Our PPO implementation, which is a modified version of a previous implementation \cite{pytorchrl}. 
\end{enumerate}

Both of these assets are under the MIT Liscence.

\end{document}